\def\year{2021}\relax
\documentclass[letterpaper]{article} 
\usepackage{aaai21}  
\usepackage{times}  
\usepackage{helvet} 
\usepackage{courier}  
\usepackage[hyphens]{url}  
\usepackage{graphicx} 
\usepackage{natbib}  
\usepackage{caption} 
\usepackage[switch]{lineno}
\usepackage{algpseudocode} 
\usepackage{algorithm} 
\usepackage{amsmath}
\usepackage{graphics}
\usepackage{microtype}
\usepackage{multirow}
\usepackage{booktabs}       
\usepackage{multicol}
\usepackage{mathrsfs}
\usepackage{url}
\usepackage{algpseudocode}
\usepackage{amssymb}
\usepackage{enumitem}
\usepackage{mathtools}
\usepackage{makecell, multirow}

\usepackage{amsfonts}       
\usepackage{amsmath}
\usepackage{multirow}
\usepackage{bm}
\usepackage{amsthm}
\usepackage{exscale}
\usepackage{relsize}
\usepackage{xr}
\newcommand{\comment}[1]{}

\newcommand{\ie}{\emph{i.e.}}
\newcommand{\eg}{\emph{e.g.}}

\title{Graph-Evolving Meta-Learning for Low-Resource Medical Dialogue Generation}

\author{

    Shuai Lin\textsuperscript{\rm 1},
    Pan Zhou\textsuperscript{\rm 2},  
    Xiaodan Liang\textsuperscript{\rm 1,3}\thanks{Corresponding Author.},
    Jianheng Tang\textsuperscript{\rm 1},
    Ruihui Zhao\textsuperscript{\rm 4}, \\
    Ziliang Chen\textsuperscript{\rm 1}, and 
    Liang Lin\textsuperscript{\rm 1,3}
    \\
}
\affiliations{

    \textsuperscript{\rm 1}Sun Yat-sen University, \textsuperscript{\rm 2}Salesforce Research, \textsuperscript{\rm 3}DarkMatter AI Inc., \textsuperscript{\rm 4}Tencent Jarvis Lab\\
    
    $\{$shuailin97, xdliang328, sqrt3tjh$\}$@gmail.com, pzhou@salesforce.com, \\
    zacharyzhao@tencent.com, c.ziliang@yahoo.com, linliang@ieee.org
}

\begin{document}
\maketitle

\begin{abstract}

	Human doctors with well-structured medical knowledge can diagnose a disease merely via a few conversations with patients about symptoms. In contrast, existing knowledge-grounded dialogue systems often require a large number of dialogue instances to learn as they fail to capture the correlations between different diseases and neglect the diagnostic experience shared among them. To address this issue, we propose a more natural and practical paradigm, i.e., low-resource medical dialogue generation, which can transfer the diagnostic experience from source diseases to target ones with a handful of data for adaptation. It is capitalized on a commonsense knowledge graph to characterize the prior disease-symptom relations. 
	Besides, we develop a Graph-Evolving Meta-Learning (GEML) framework that learns to evolve the commonsense graph for reasoning disease-symptom correlations in a new disease, which effectively alleviates the needs of a large number of dialogues. More importantly, by dynamically evolving disease-symptom graphs, GEML also well addresses the real-world challenges that the disease-symptom correlations of each disease may vary or evolve along with more diagnostic cases. Extensive experiment results on the CMDD dataset and our newly-collected Chunyu dataset testify the superiority of our approach  over state-of-the-art approaches. 
	Besides, our GEML can generate an enriched dialogue-sensitive knowledge graph in an online manner, which could benefit other tasks grounded on knowledge graph.
\end{abstract}

\section{Introduction} \label{sec:intro}

    \noindent 	Medical dialogue system (MDS) aims to converse with patients to inquire additional symptoms beyond their self-reports and make a diagnosis automatically, which has gained increasing attention \cite{lin-etal-2019-enhancing,wei2018task,xu2019end}.
	It has a significant potential to simplify the diagnostic process and relieve the cost of collecting information from patients \cite{kao2018context}. Moreover, preliminary diagnosis reports generated by MDS may assist doctors to make a diagnosis more efficiently.  
	Because of these considerable benefits, many researchers devote substantial efforts  to address critical sub-problems in  MDS, such as natural language understanding ~\cite{shi2020understanding,lin-etal-2019-enhancing}, dialogue policy learning~\cite{wei2018task}, dialogue management~\cite{xu2019end}, and make promising  progress to build a satisfactory MDS. 

	\begin{figure}[t]
	\centering
	\includegraphics[width=0.98\linewidth]{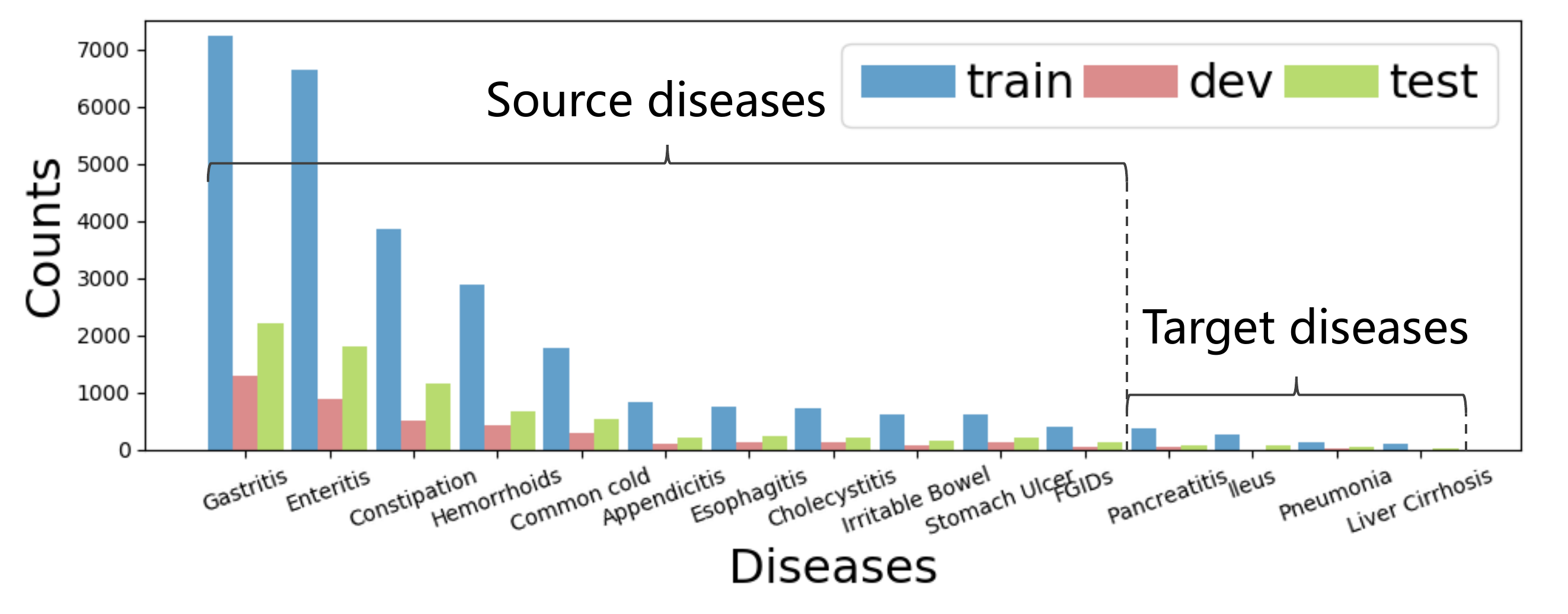}
	\caption{Statistics  of 15 diseases in our newly collected \emph{Chunyu} dataset from the real world. One can observe the notable data-imbalance phenomenon of different diseases. Thus it is highly desirable to study how to transfer the diagnostic experience among diseases.}

	\label{fig:data split}
	\end{figure}
	
	Medical dialogue generation (MDG), which generates responses in natural language to request additional symptoms or make a diagnosis, is critical in MDS but rarely studied.
	Conventional generative dialogue models often employ neural sequence modeling~\cite{sutskever2014sequence,vaswani2017attention} and cannot be applied to the medical dialogue scenario directly in absence of medical knowledge. Recently, large-scale pre-training language models \cite{devlin2018bert,radford2019language,song2019mass} over unsupervised corpora have achieved significant success. 
    However, fine-tuning such large language models in the medical domain requires sufficient task-specific data ~\cite{bansal2019learning,dou-etal-2019-investigating} so as to learn the correlations between diseases and symptoms.
	Unfortunately, as depicted in Fig.~\ref{fig:data split}, there are a large portion of diseases that only have a few instances in practice, which means that newly-coming diseases in the realistic diagnosis scenario are often under low-resource conditions. Therefore, it is highly desirable to transfer the diagnostic experience from high-resource diseases to others of data scarcity. 
	Besides, existing knowledge-grounded approaches~\cite{liu-etal-2018-knowledge,lian2019learning} may fail to perform such transfer well, as they only learn one unified model for all diseases and ignore the specificity and relationships of different diseases.
	Finally, in practice, the disease-symptom relations of each disease may vary or evolve along with more cases, which is also not considered in prior works.  

	\textbf{Contributions.}	To address the above challenges, we first propose an end-to-end  dialogue system for the low-resource medical dialogue generation.
	This model integrates three components seamlessly, a hierarchical context encoder,  a meta-knowledge graph reasoning (MGR) network and a graph-guided response generator. Among them, the context encoder encodes  the conversation into hierarchical representations. For MGR, it mainly contains a parameterized meta-knowledge graph, which is initialized by a prior commonsense graph and characterizes the correlations among diseases and symptoms.  When fed into the context information, MGR can adaptively evolve its meta-knowledge graph to reason the disease-symptom correlations and then predict related symptoms of the patient in the next response to further determine the disease. Finally, the response generator generates a response for symptoms request  under the guidance of the meta-knowledge graph. 

	The second contribution is that we further develop a novel Graph-Evolving Meta-Learning (GEML) framework to  transfer the diagnostic experience in the low-resource scenario. Firstly, GEML trains the above medical dialogue model under the meta-learning framework. It regards generating responses to a handful of dialogues as a task and learns a meta-initialization for the above dialogue model that can fast adapt to each task of the new disease with limited dialogues. In this way, the learnt model initialization contains sufficient meta-knowledge\footnote{We name such knowledge as ``meta-knowledge" since it is obtained through meta-training from different source diseases.} from all source diseases and can serve as a good model initialization to quickly transfer meta-knowledge to a new disease. More importantly, GEML also learns a good parameterized meta-knowledge  graph in the MGR module to characterize the disease-symptom relationships from source diseases. Concretely, under the meta learning framework, for each disease, GEML enriches the meta-knowledge graph via constructing a global-symptom graph from the online dialogue examples. In this way, the learnt meta-knowledge graph can bridge the gap between the commonsense medical graph and the real diagnostic dialogues and thus can be fast evolved for the new target disease. Thanks to graph evolving, the dialogue model can request patients for underlying symptoms more efficiently and thus improve the diagnostic accuracy. Besides,  GEML can also well address the real-world challenge that the disease-symptom correlations could vary along with more cases, since the meta-knowledge  graph is trainable based on collected dialogue examples.

	Finally, we  construct a large medical dialogue dataset, called \emph{Chunyu}\footnote{Code and dataset are released at https://github.com/ha-lins/GEML-MDG.}. 
	It covers 15  kinds of diseases and 12,842 dialogue examples totally, and  is much larger than the existing CMDD  medical dialogue dataset~\cite{lin-etal-2019-enhancing}. The more challenging benchmark can better comprehensively evaluate the performance of medical dialogue systems.  Extensive experimental results on both datasets demonstrate the superiority of our method over the state-of-the-arts.

\section{Related Work}
\textbf{Medical Dialogue System (MDS). } 
	Recent research on MDS  mostly focus on the natural language understanding (NLU) or dialogue management (DM) with the line of pipeline-based dialogue system. Various NLU problems have been studied to improve the MDS performance, \eg, ~entity inference \cite{du-etal-2019-learning, lin-etal-2019-enhancing, liu2020meddg}, symptom extraction \cite{du-etal-2019-extracting} and slot-filling \cite{shi2020understanding}. For medical dialogue management, most works~\cite{dhingra2017towards,li-etal-2017-end} focus on reinforcement learning (RL)  based task-oriented dialogue system. 
	\citet{wei2018task} proposed to learn dialogue policy with RL to facilitate automatic diagnosis. 
	\citet{xu2019end} incorporated the knowledge inference into dialogue management via RL. However,  no attention  has been paid  to the medical dialogue generation, which is a critical recipe in MDS. Differing from existing approaches, we investigate to build an end-to-end graph-guided medical dialogue generation model directly.
	
    \noindent\textbf{Knowledge-grounded Dialog Generation.}  Recently, dialogue generation grounded on extra knowledge is emerging as an important step towards human-like conversational AI, where the knowledge could be derived from or open-domain knowledge graphs \cite{zhou2018commonsense,zhang2020grounded,moon2019opendialkg} or retrieved from unstructured documents \cite{lian2019learning,zhao2019low,kim2020sequential}. Different from them, our MDG model is built on the dedicated medical-domain knowledge graph and further require evolving it to satisfy the need for the real-world diagnosis.
	
	\noindent\textbf{Meta-Learning. }  By meta-training a model initialization from training tasks with the ability of fast adaptation to new tasks,  meta-learning~\cite{pmlr-v70-finn17a,zhou2019metalearning,zhou2020task} has achieved promising results in many NLP areas, such as machine translation \cite{gu-etal-2018-meta}, task-oriented dialogues \cite{qian-yu-2019-domain,mi2019meta}, and text classification \shortcite{obamuyide-vlachos-2019-model,wu-etal-2019-learning}.  But there is the few effort to devote meta-learning into MDS, which requires grounding on the external medical knowledge and reasoning for disease-symptom correlations. In this work, we employ the Reptile~\cite{nichol2018first}, one first-order model-agnostic meta learning approach,  because of its efficiency and effectiveness, and enhance it with the meta-knowledge graph reasoning and evolving.

    \section{Task Definition: Low-Resource MDG}
    Grounded on the external medical knowledge graph $A$, the medical dialogue generation models take the dialogue context $U = \{u_1, \dots, u_{t-1}\}$ as input and aim to (1) generate the next response $R = u_t$ and (2) predict the disease or symptom entity $E = e_t$ appearing in the next response as:
    \begin{equation}
    f_{\theta}(R, E | U, A ; \theta) = p\left( u_{t}, e_{t} | u_{1:t}, A; \theta\right),
    \label{eq:obj-mdg}
    \end{equation}
    
    Given the abundant dialogue examples of $K$ different source diseases $S_k$, the task of low-resource MDG require to obtain a good model initialization during meta-training process: 
    
    \begin{equation}
    \theta_{\rm meta}: (U, A) \times S_k \rightarrow (R_{source}, E).
    \label{eq:source-model}
    \end{equation}
        
    For the adaptation to the new target disease $T$, we fine-tune the model $\theta_{\rm meta}$  with minimal dialogue examples (e.g., 1\% $\sim$ 10\% of the source disease) and require the induced model $\theta_{\rm target}$ to perform well in the target disease:
    
    \begin{equation}
    \theta_{\rm target}: (U, A) \times T \rightarrow (R_{target}, E).
    \label{eq:target-model}
    \end{equation}

    \section{End-to-End Medical Dialogue Model}\label{endtoendmodel}
\begin{figure*}[t]
		\centering
		\includegraphics[width=1\linewidth]{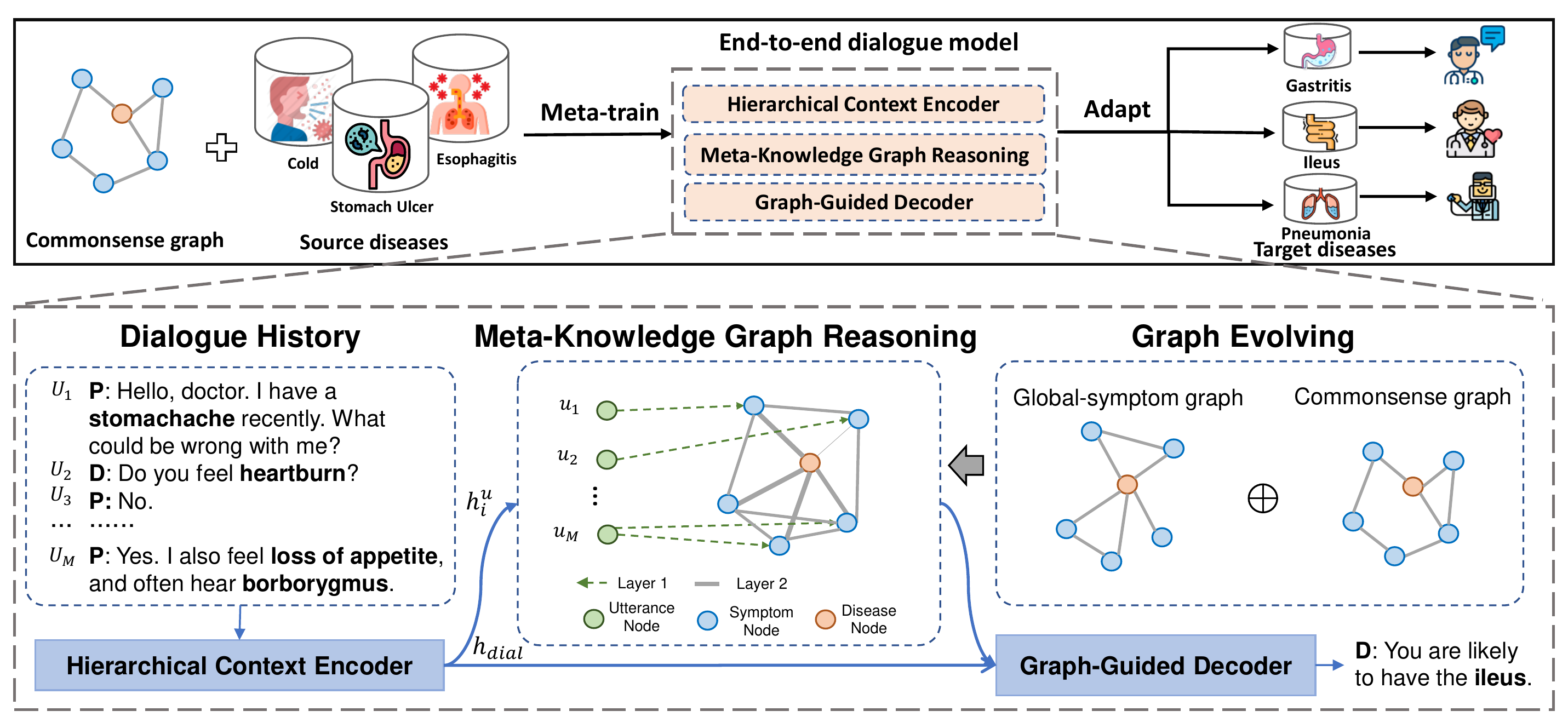}
		\caption{Framework overview. \textbf{Upper:} The overview of the proposed GEML-MGR for the low-resource medical dialogue generation. The GEML-MGR first goes through the meta-training phase to learn and evolve a meta-knowledge graph and then adapts to new target diseases. \textbf{Lower:} The architecture of the end-to-end medical dialogue model, which integrates three components seamlessly: hierarchical context encoder, meta-knowledge graph reasoning and graph-guided response generator. }
		\label{fig:framework}
	\end{figure*}

    In this section, we elaborate our end-to-end dialogue model whose framework is illustrated in Fig.~\ref{fig:framework}. The proposed approach integrates three components seamlessly, including hierarchical context encoder, meta-knowledge graph reasoning (MGR) and graph-guided response generator. Concretely, the context encoder  first encodes the conversation history into hierarchical context representations. 
    Then MGR incorporates the obtained representations into the knowledge graph reasoning process for the comprehension of the disease-symptom correlations. 
    Finally, the graph-guided decoder generates informative responses  via a well-designed copy mechanism over graph entity nodes. We will introduce them in turn. 
	
	\subsection{Hierarchical Context Encoder}

	We first utilize a hierarchical context encoder \cite{serban2016building} to encode the dialogue history and obtain the hierarchical hidden representations of the context. Formally, given a dialogue context $U=(u_1,\ldots,u_l)$, the hierarchical context encoder first exploits a long short term memory (LSTM) network  to encode each utterance into a hidden representation:
	\begin{equation}
	\bm{h}^u_{i} = \mathtt{LSTM}_{\theta_u}(\bm{e}^{i}_{1}, \ldots, \bm{e}^i_{j},\ldots, \bm{e}^{i}_{l_i}),\label{lstm}
	\end{equation}
	where $\bm{e}^{i}_{j}$ is the embedding of the  $j$-th token in the $i$-th utterance. Then these hidden representations $\{\bm{h}^u_{i}, i=1,\cdots,l\}$ of utterances are fed into another LSTM to obtain the representation of the entire dialogue history as
	\begin{equation}
	\bm{h}_{\mbox{\tiny{dial}}} = \mathtt{LSTM}_{\theta_d}(\bm{h}^{u}_{1}, \ldots, \bm{h}^u_{j},\ldots, \bm{h}^{u}_{l}).
	\end{equation}
	After obtaining utterance-level and dialogue-level representations, as shown in Fig.~\ref{fig:framework}, we use $\bm{h}^u_{i}$  to initialize the utterance node features of the knowledge graph in Sec.~\ref{subsec:MGR}, and then adopt $\bm{h}_{\mbox{\tiny{dial}}}$ as the initial state  of the decoder LSTM in Sec.~\ref{section:CopyNet}.  
	
	\subsection{Meta-Knowledge Graph Reasoning} \label{subsec:MGR}
	Based on the obtained utterance representation $\bm{h}^u_{i}$, we need to learn the disease-symptom correlations and further inquiry the patient the existence of related symptoms to verify. To this end, we devise a meta-knowledge graph reasoning  (MGR) network to learn and reason the above correlations.   
	In practice, one often has a prior commonsense disease-symptom graph which  roughly contains such correlations,  \eg, ~\emph{cold} being indicated by the symptom \emph{cough}. Our MGR aims to (1) reason the correlations over diseases and symptoms through conversations with patients, (2) predict the possible symptoms in the next inquiry/response to the patient, (3) evolve this commonsense disease-symptom graph into a meta-knowledge graph with the graph evolving meta-learning (GEML) framework. Here we focus on introducing the first two points and present our GEML in Sec.~\ref{sec:GEML}.
	
	In practice, the commonsense disease-symptom graph can be derived from the Chinese Symptom Library in OpenKG\footnote{OpenKG is a Chinese open knowledge graph project. The library is available at http://openkg.cn/dataset/symptom-in-chinese.}. The library contains a huge amount of triples, \eg, ~(\emph{Diarrhea}, \emph{related symptom},  \emph{Gastroenteritis}). Formally, we denote the commonsense graph as $\mathcal{G}=(\mathcal{V}^e, \mathcal{A}, \mathcal{X})$, where $\mathcal{V}^e=\{{v_{1}^e, \ldots, v^e_{m}}\}$ is the set of entity nodes, $\mathcal{A}$ is the corresponding adjacency matrix and $\mathcal{X} \in  \mathcal{R}^{|\mathcal{V}^e| \times F} $ is the node feature matrix (${F}$ is the number of features in each node). In the graph $\mathcal{G}$, each entity node $v_{i}^e \in \mathcal{V}^e$ denotes a symptom or disease. The feature vector of each entity node, i.e., each row of the feature matrix $\mathcal{X}$, is trainable. Besides, we have a utterance node set denoted as $\mathcal{V}^u=\{{v_{1}^u, \ldots, v_{l}^u}\}$, where the input feature of each utterance node ${v_{i}^u}$ is initialized by the  representation $\bm{h}^u_{i}$ obtained in Eqn.~\eqref{lstm}. To incorporate the context information into the knowledge graph reasoning, we connect each utterance node with all entity nodes it included.

	Now we introduce the \textit{graph reasoning} process over diseases and symptoms. To boost the information propagation among entity nodes, we build a meta-knowledge graph  where each entity node indicates a disease or symptom. Inspired by the graph attention network \cite{velickovic2018graph}, we devise the meta-knowledge graph reasoning (MGR) network that consists of two graph reasoning layers. In the first layer, entity nodes that occur in the dialogue history are activated by aggregating the information from their corresponding utterance nodes. Then in the second layer, these activated entity nodes diffuse the information to their neighborhood nodes for the correlation reasoning. Next we present the single graph reasoning layer that used to construct the MGR (through stacking this layer).  Let $\mathcal{N}_i$  be the neighbor set of the node $i$ according to the adjacency matrix $\mathcal{X}$.  With the input feature  ${h}^e_j$ of some neighborhood nodes $j \in \mathcal{N}_i$, the  graph reasoning layer updates the representation of node $i$ as:  
	\begin{align}\label{eqnatt}
	\begin{gathered}
	{h}^e_i = \sigma\big(\mathsmaller{\sum}\nolimits_{j\in\mathcal{N}_i} \alpha_{ij} {\bm W_0}{h}^e_j\big) \\
	\alpha_{ij} = \mathrm{softmax}_j(e_{ij})=\exp(e_{ij})/\mathsmaller{\sum}\nolimits_{k\in\mathcal{N}_i} \exp(e_{ik})
	\end{gathered}
	\end{align}
	where ${\bm W_0}\in\mathcal{R}^{F \times F}$ is a weight matrix and $e_{ij}$ is the attention coefficient that indicates the \emph{importance} of entity node $j$ to node $i$. Following ~\cite{bahdanau2014neural}, the attention coefficient $e_{ij}$ is computed as
	\begin{equation}
	e_{ij}=~\mathrm{Sigmoid}(\bm  a^T {\bm W}_1 [ {h}^e_i|| {h}^e_j]),
	\end{equation}
	where $\bm a\in\mathcal{R}^{H \times 1}$ is a trainable vector, ${\bm W}_1\in\mathcal{R}^{H \times 2F} $ is a weight matrix and $||$ indicates the concatenation. Note we inject the graph structure (i.e. the adjacency matrix $\mathcal{A}$) into the graph reasoning layer as we only compute the $e_{ij}$ for neighborhood nodes $j$ of $i$. In Sec. \ref{subsec:meta-graph-evolving}, we will elaborate how to evolve the meta-knowledge graph structure in a meta-learning paradigm. By stacking two graph reasoning layers, each entity node could grasp enough information from other related nodes. As shown in Fig.~\ref{fig:framework}, we then feed the final entity node representations$\{{h}^e_i, i=1,\cdots,m\}$ into the response generator to infer possible entities in the next-turn response. To this end, we introduce the \emph{entity prediction} task beyond response generation. Concretely, we feed the final node representations $\{{h}^e_i, i=1,\cdots,m\}$ into a feed-forward layer and predict possible entities in the next response via the binary classification over all graph entity nodes. In this way, our MGR network can mine and reason the disease-symptom correlations,  and thus predict underlying entities in the next response to diagnose more accurately.

	\subsection{Graph-guided Response Generator} \label{section:CopyNet}

    To incorporate the knowledge graph into the generation, we devise a graph-guided response generator with a copy mechanism adapted from \cite{see-etal-2017-get}. The main modification is that we apply the copy mechanism over the graph nodes distribution instead of the input source. 
	More concretely, under the guidance of the entity node representations $\{{h}^e_i, i\!=\!1,\cdots\!,m\}$, the decoder generates each word at the time step $t$ via sampling from the vocabulary or copying directly from the graph entity nodes set $E$ as:
	\begin{equation}
	\small
	\bm{P}^{(t)}_{out} =  ~ g_t \cdot \bm{P}^{(t)}_{V} + (1 - g_t) \cdot \bm{P}^{(t)}_{\bm{E}},
	\label{eq:p-out}
	\end{equation}
	where $\bm{P}^{(t)}_{V}$ is the normal vocabulary distribution from the decoder LSTM; $\bm{P}^{(t)}_{\bm{E}}$ is the attention distribution over graph entity nodes. The soft switch  $g_t \in [0,1]$ to choose between sampling or copying is calculated given the decoder input $x_t$ and the decoder state $s_t$ as:
	\begin{equation}\label{eq:g-t}
	g_t =  ~ \sigma(\bm W_{2} \cdot [\bm x_t;\bm s_t;\bm h^a_t])\quad\text{with}\quad 
	\bm h^a_t =  ~ \mathsmaller{\sum}\nolimits_i \alpha^e_i \cdot {h}^e_i, 
	\end{equation}
	where $\bm W_{2}$ is a trainable matrix and $\sigma$ is the $\mathrm{Sigmoid}$ function.  The aggregation vector $h^a_t$ is computed through the weighted sum over the node representations ${h}^e_i$, and $\alpha^e_i$ is the attention weight calculated as ~\cite{bahdanau2014neural}. With the above graph-guided copy mechanism, the response generator can achieve the more accurate symptom request and disease diagnosis result.

	\section{Graph-Evolving Meta-Learning}\label{sec:GEML}
	%
	In this section, we present a Graph-Evolving Meta-Learning (GEML) framework which helps the above end-to-end medical dialogue model to handle the low-resource setting. This setting is more practical and challenging since many diseases in the real world are rare and costly to annotate as mentioned in Sec. \ref{sec:intro}. To address this challenge, GEML uses meta-knowledge transfer and meta-knowledge graph evolving to transfer the diagnostic experience across different diseases. We'll introduce them in turn.

	\subsection{Meta-Knowledge Transfer}
	The methodology of meta-knowledge transfer is to meta-train an  end-to-end medical dialogue model $f_{\theta_{\rm meta}}$ parameterized by $\theta_{\rm meta}$ with a fast adaptation capacity to new diseases with only limited data.  To this end, we follow the meta-learning framework and use existing  dialogue data of $k$ source diseases to create a task set $\mathscr{T} \!=\! \{\{\mathcal{T}_{i}^{1}\}_{i=1}^{N_1},  \{\mathcal{T}_{i}^{2}\}_{i=1}^{N_2} \dots,\{\mathcal{T}_{i}^{k}\}_{i=1}^{N_k}\}$, where each task $\mathcal{T}_{i}^{k}$ represents generating responses to a handful of dialogues in the $k$-th disease. Each task $\mathcal{T}^k_{i}  \in \mathscr{T} $ has only a few dialogue samples, which can be further split into the training (support) set $\mathcal{D}_{tr}^{\mathcal{T}_{i}}$ and the validation (query) set $\mathcal{D}_{va}^{\mathcal{T}_{i}}$.
	Then in the meta-training stage, given a  model initialization $\theta_{\rm meta}$, we require that  $\theta_{\rm meta}$ can  fast adapt to  any task $\mathcal{T}_{i}  \in \mathscr{T} $ through one gradient update:
	\begin{equation}\label{eq:inner-update}
	\theta_{i}\hspace{-0.1em}=\hspace{-0.1em}\theta_{\rm meta}\hspace{-0.1em}-\hspace{-0.1em}\beta \nabla_{\theta} \mathcal{L}_{\mathcal{D}_{tr}^{\mathcal{T}_{i}}}\left(f_{\theta_{\rm meta}}\right),
	\end{equation}
	where  $\mathcal{L}_{\mathcal{D}_{tr}^{\mathcal{T}_{i}}}$ is the training loss function of task $\mathcal{T}_{i}$ and $\beta$ denotes a learning rate.  
	To measure the quality of the adapted parameter $\theta_{i}$, MAML \cite{pmlr-v70-finn17a}, which is an optimization-based meta-learning approach, requires  $\theta_{i}$ to have small validation loss on the validation set $\mathcal{D}_{va}^{\mathcal{T}_{i}}$. In this way, it can compute the gradient of validation loss and update the initialization  $\theta_{\rm meta}$ as 
	\begin{equation}\label{Afafdcsa}
	\theta_{\rm meta}\hspace{-0.1em}=	\theta_{\rm meta} - \gamma \nabla  \mathcal{L}_{\mathcal{D}_{va}^{\mathcal{T}_{i}}}\big( \theta_{\rm meta}\hspace{-0.1em}-\hspace{-0.1em}\beta \nabla_{\theta} \mathcal{L}_{\mathcal{D}_{tr}^{\mathcal{T}_{i}}}\left(f_{\theta_{\rm meta}}\right) \big),
	\end{equation}
where  $\gamma$ is step size. To alleviate the computational cost for the second-order gradient, \ie~Hessian matrix, in Eqn.~\eqref{Afafdcsa},  Reptile~\cite{nichol2018first} approximates the second derivatives of the validation loss  as
	\begin{equation}
	\theta_{\rm meta} \leftarrow \theta_{\rm meta} + \gamma \mathsmaller{\frac{1}{|\{\mathcal{T}_i\}|} \sum}\nolimits_{\mathcal{T}_i \sim p(\mathcal{T})} (\theta_i - \theta_{\rm meta}).
	\end{equation}
In this work, we use Reptile  to update the initialization $ \theta_{\rm meta}$ because of its effectiveness and efficiency. After obtaining the initialization $ \theta_{\rm meta}$, given a new target disease with only a few training data $\mathcal{D}_{tr}$, we can adapt the model $f_{\theta_{\rm meta}}$ with initialization $ \theta_{\rm meta}$  to the disease quickly via a few gradient steps to obtain the disease-adapted parameters. This fast adaptation ability comes from that, in the meta-training phase, we have already simulated the fast learning to a new disease via few steps of gradient descent on few validation data.

Note this meta-knowledge transfer only considers the fast adaptation in terms of model parameters and ignores the flaw of the sparsity in the commonsense graph. To address this problem,  we devise a graph evolving approach to evolve the commonsense graph such that it can be tailored to the current disease and integrated with the dialogue instances better.

	\subsection{Meta-Knowledge Graph  Evolving} \label{subsec:meta-graph-evolving}
	
	Since the commonsense graph is sparse and does not cover enough symptom entities,  there is a gap between this prior  graph and the real dialogue examples. For instance, ``dysbacteriosis" may appear in the consultation of a patient while it doesn't exist in the commonsense graph since it is comparatively rare. 
	To address the challenge, we propose to evolve the commonsense graph capitalized on the dialogue instances and learn the induced meta-knowledge graph during the meta-training and adaptation phases. 
	Inspired by \citet{lin-etal-2019-enhancing} that shows the related symptom entities have a certain probability of co-occurrence in the same dialogue, we construct a global-symptom graph $ \mathcal{G^*}=(\mathcal{V^*}, \mathcal{A^*}, \mathcal{X^*})$, 
	where $\mathcal{V^*}=\{v_1,\ldots,v_n\}$ is the set of nodes, $\mathcal{A^*}$ is the corresponding adjacency matrix and $\mathcal{X^*} \in  \mathbb{R}^{|\mathcal{V}^*| \times N} $ is the node feature matrix. Concretely, the proposed approach first collects all observed dialogue examples in an online manner. Then if two entities co-occur in a dialogue example, there is an edge between both nodes in $\mathcal{A^*}$. 
	The meta-knowledge graph is initialized with the adjacency matrix $\mathcal{A}$ of the prior commonsense graph and updated as:
	\begin{equation}
	\mathcal{A_{\rm meta}} = \mathcal{A} \oplus \mathcal{A^*},
	\end{equation}
	where $\oplus$ denotes the element-scale logic operator \texttt{OR}.  In this way, updating the adjacency matrix  $\mathcal{A_{\rm meta}} $ can reason the existence of edges among entity nodes. The structure of $\mathcal{A^*}$ is dynamically evolved along with more dialogue cases, which leads to the enrichment of the meta-knowledge graph synchronously, i.e., adding more nodes and edges. 
	
	The above approach for graph structure evolving can infer the existence of disease-symptom correlations while ignoring its intensity. To characterize such relations more delicately, GEML further learns the weight values of the meta-knowledge graph $\mathcal{A_{\rm meta}}$ with Eqn. (6) during the meta-training and adaptation phases (while not during the testing). Finally, GEML utilizes the cross-entropy loss of the entity prediction task (in Sec.~\ref{subsec:MGR}) to guide the learning of $\mathcal{A_{\rm meta}}$ efficiently, and we denote it as $\mathcal{L}_e$. 
	
\begin{table}[t]
\small
\centering
\scalebox{0.9}{
\begin{tabular}{rcc}
	\cmidrule[\heavyrulewidth]{1-3}
	\textbf{Dataset}   & \textbf{CMDD} &\textbf{ Chunyu}  \\
	\cmidrule{1-3}
	\# Disease types               & 4 & 15    \\ 
	\# Dialogues                   & 2067   & 12,842   \\ 
	{\# Utterances per dialogue} & 42.09  & 24.7 \\ 
	{\# Entities per dialogue}  & 7.5  & 12.9\\ 
	{\# Words per utterance}    & 10.0  & 10.6\\ 
	\cmidrule[\heavyrulewidth]{1-3}
\end{tabular}}
	\caption{Statistics of the CMDD dataset ~\cite{lin-etal-2019-enhancing} and our Chunyu dataset.}
\label{tab:dataset}
\end{table}

	\subsection{End-to-End Training Loss} 
	In this section, we'll introduce the loss function for each task, i.e.,  $\mathcal{L}_{\mathcal{D}_{tr}^{\mathcal{T}_{i}}}$ in Eqn.~\eqref{eq:inner-update} in detail. 
	The generation loss function is the negative log-likelihood of generating the response $R=\{r_1,\ldots,r_m\}$ given the input dialogue context $U=\{u_1,\ldots,u_n\}$ as:
	\begin{equation}
	\mathcal{L}_g = -\mathsmaller{\frac{1}{|R| }\sum}\nolimits^{|R|}_{i=1}      \text{ log}p(r_i|U;\theta_{\rm meta}).
	\end{equation}
	The final training objective couples $\mathcal{L}_g$ with $\mathcal{L}_e$ and is of :
	\begin{equation}
	\mathcal{L} = \mathcal{L}_g + \lambda\mathcal{L}_e,
	\end{equation}
	where the constant $\lambda$  balances  the loss  $\mathcal{L}_g $  and the entity prediction loss $\mathcal{L}_e$ in Sec.~\ref{subsec:meta-graph-evolving}.
	
\begin{table*}[!t]
		\scriptsize
		\centering
		\scalebox{0.9}{
		\begin{tabular}{l|l|cc|cc|cc|cc|cc|cc}
			\cmidrule[\heavyrulewidth]{1-14}
			\textbf{\multirow{3}{*}{Dataset}}& \textbf{\multirow{3}{*}{Method}} & 
			\multicolumn{10}{c|}{\textbf{Automatic Metrics}}
			& \multicolumn{2}{c}{\textbf{Human Evaluation}} \\
			\cline{3-4}\cline{5-6}\cline{7-8}\cline{9-10}\cline{11-12}\cline{13-14} 
			& &
			\multicolumn{2}{c|}{\textbf{Target Disease 1}}&
			\multicolumn{2}{c|}{\textbf{Target Disease 2}}&
			\multicolumn{2}{c|}{\textbf{Target Disease 3}}& 
			\multicolumn{2}{c|}{\textbf{Target Disease 4}}&
			\multicolumn{2}{c|}{\textbf{Average}} & \textbf{Knowledge} & \textbf{Generation}\\
	
			& & \textbf{BLEU}   & \textbf{Enti.-F1}  
			& \textbf{BLEU}  & \textbf{Enti.-F1} 
			& \textbf{BLEU}   & \textbf{Enti.-F1} 
			& \textbf{BLEU}   & \textbf{Enti.-F1} 
			& \textbf{BLEU}   & \textbf{Enti.-F1} 
			& \textbf{Rationality} & \textbf{Quality}\\
			\cmidrule[\heavyrulewidth]{1-14}
	\textbf{\multirow{11}{*}{CMDD}}

	& PT-NKD\shortcite{liu-etal-2018-knowledge}   & 39.16 & 25.71 & 30.1 & 16.39 & 32.02 & 12.5 & 30.64 & 22.22 & 32.98 & 19.21 & 2.38 & 2.5  \\
	& PT-POKS\shortcite{lian2019learning}  & 41.51 & 33.33 & 12.7 & 35.88 & 33.27 & 22.53 & 31.06 & 27.45 & 29.63 & 29.79 & 2.87 & 2.92 \\
	& PT-MGR   & 43.96 & 36.36 & 31.31 & 18.19 & 37.78 & 23.89 & 31.95 & 28.33 & 36.25 & 26.69 & 3.26 & 3.28 \\\cmidrule{2-14}
	
	& FT-NKD\shortcite{liu-etal-2018-knowledge}  & 42.72 & 28.57 & 30.32 & 31.74 & 34.67 & 21.69 & 32.45 & 31.58 & 35.04 & 28.4 & 3.13 & 3.42 \\
	& FT-POKS\shortcite{lian2019learning} & 41.8 & 42.78 & 32.25 & 37.15 & 35.36 & 25 & 32.56 & 25.97 & 35.49 & 32.72 & 3.06 & 2.94 \\
	& FT-MGR  & 45.23 & 45.78 & 36.18 & 38.81 & 34.59 & 23.08 & 33.07 & 29.36 & 37.27 & 34.26 & 3.39 & 3.56 \\\cmidrule{2-14}
	
	& Meta-NKD   & 41.23 & 42.5 & 32.5 & 32.27 & 34.61 & 24.17 & 32.28 & 30.17 & 35.16 & 32.29 & 3.11 & 3.39 \\
	& Meta-POKS  & 40.64 & 35.14 & 34.25 & 37.14 & 36.84 & 27.69 & 33.7 & 28.25 & 36.35 & 32.06 & 3.31 & 3.3 \\
	& Meta-MGR   & 45.78 & 40.03 & \bf{35.51} & \bf{47.27} & 37.01 & 25.14 & 33.71 & 31.02 & 38 & 35.87 & 3.56 & 3.65 \\\cmidrule{2-14}
	
	& GEML-MGR     & \bf{48.39} & \bf{45.7} & 33.59 & 46.37 & \bf{43.88} & \bf{28.57} & \bf{37.07} & \bf{36.84} & \bf{40.73} & \bf{39.37} & \bf{3.72} & \bf{3.77} \\
	\cmidrule[\heavyrulewidth]{1-14}

	\textbf{\multirow{11}{*}{Chunyu}}
	& PT-NKD\shortcite{liu-etal-2018-knowledge} & 16.01 & 4.54 & 20.75 & 18.75 & 13.17 & 11.27 & 17.45 & 17.54 & 16.84 & 13.03 & 2.78 & 3.26 \\
	& PT-POKS\shortcite{lian2019learning}   & 15.24 & 10.13 & 21.34 & 18.46 & 15.25 & 12.99 & 18.9 & 21.65 & 17.68 & 15.82 & 2.81 & 2.93 \\
	& PT-MGR    & 15.5 & 7.5 & 25.42 & 21.91 & 18.13 & 13.95 & 19.46 & 29.76 & 19.62 & 25.63 & 3.11 & 3.29 \\\cmidrule{2-14}

	& FT-NKD\shortcite{liu-etal-2018-knowledge} & 16.35 & 4.87 & 19.52 & 20.68 & 15.28 & 18.18 & 18.49 & 26.74 & 17.41 & 17.62 & 2.89 & 3.37 \\
	& FT-POKS\shortcite{lian2019learning} & 14.46 & 17.24 & 21.63 & 32.16 & 16.45 & 23.08 & 18.18 & 27.32 & 17.68 & 24.95 & 3.12 & 2.96\\
	& FT-MGR  & 18.38 & 28.57 & 25.61 & 38.88 & 19.53 & 22.27 & 20.41 & 30.15 & 20.98 & 29.97 & 3.29 & 3.17 \\\cmidrule{2-14}
	
	& Meta-NKD   & 17.28 & 33.96 & 22.20 & 41.31 & 17.54 & 18.2 & 22.71 & 32.39 & 19.93 & 31.47 & 3.18 & 3.34\\
	& Meta-POKS  & 17.87 & 23.18 & 24.76 & 42.86 & 16.46 & 23.35 & 16.71 & 22.22 & 18.96 & 27.9 & 3.12 & 3.19\\
	& Meta-MGR   & 19.65 & 32.12 & 26.43 & 42.35 & \bf 19.74 & 27.16 & 21.32 & 32.45 & 21.79 & 33.52 & 3.41 & 3.38 \\ \cmidrule{2-14}
	
	& GEML-MGR  & \bf {22.53} & \bf{34.72} & \bf{26.74} & \bf{47.88} & 19.13 & \bf{36.96} & \bf{23.89} & \bf{35.13} & \bf{23.07} & \bf{38.67} & \bf {3.52} & \bf {3.48} \\
	\cmidrule[\heavyrulewidth]{1-14}

		\end{tabular}
		}
		\caption{ Results on the two datasets in terms of automatic metrics ($\times 10^2$) and human evaluation (on a 5-point scale). {\bf Top:} For the CMDD dataset, target diseases from 1 to 4 refer to “bronchitis”, “functional dyspepsia”, “infantile diarrhea” and “upper respiratory infection” respectively. {\bf Bottom:} For the Chunyu dataset, target diseases from 1 to 4 refer to ``liver cirrhosis", ``ileus", ``pneumonia", and ``pancreatitis". }
	\label{tab:cmdd}
	\end{table*}

\section{Experiments}
	
Here  we conduct extensive experiments on the CMDD dataset \cite{lin-etal-2019-enhancing} and the newly-collected Chunyu dataset to demonstrate the benefits of GEML. 

	\noindent\textbf{Datasets. }  The CMDD dataset \cite{lin-etal-2019-enhancing} has 2,067 conversations totally ranging 4 pediatric diseases with approximately equal counts, while neglects the data-imbalance problem among diseases. 
	To pose the challenge, we collect a much larger medical dialogue dataset, namely Chunyu, which contains 15 diseases with comparatively distinct data ratios. As depicted in Fig.~\ref{fig:data split}, the counts of each disease in Chunyu are significantly various and thus we can treat four low-resource diseases as target ones. The data statistics of two datasets are depicted in Table \ref{tab:dataset}. The raw data of Chunyu is obtained from the Gastroenterology department of the Chinese online health community \emph{Chunyu}\footnote{https://www.chunyuyisheng.com/}. It contains 15 gastrointestinal diseases and 62 symptoms in total.	We use hand-crafted rules provided by doctors to label entities for each instance. The instances with very few turns or entities and with private information have been all discarded.

\noindent\textbf{Experimental Settings. } To apply the meta-learning, we consider generating responses to a handful of dialogues in one disease as a task. For Chunyu, as shown in Fig.\ref{fig:data split}, high-resource diseases with more than 500 training instances are treated as source diseases and the remaining four low-resource ones as target diseases, whose size of adaptation data is ranging from 80 $\sim$ 200. For CMDD, we adopt the standard \emph{leave-one-out} setup, i.e., using four diseases for meta-training and the one target disease left for adaptation (with the data size of 150 dialogues). 

All experiments are based on the AllenNLP toolkit~\cite{gardner2018allennlp}.  We implement encoders and decoders with a single-layer LSTM ~\cite{hochreiter1997long}, and use pkuseg~\cite{pkuseg} toolkit to segment   Chinese words. We set both dimensions of the hidden state and word embedding to 300 for LSTM. Adam optimization is adopted with the initial learning rate of 0.005 and the mini-batch size of 16. The maximum training epochs are set to 100 with the patience epoch of 10 for early-stopping. 
The best hyper-parameter $\bm{\lambda}$ to balance the generation loss and the entity loss is 8. All baselines share the same configuration settings.

\noindent\textbf{Baselines. } 
We first  compare our  base dialogue model MGR with two   knowledge-grounded dialogue systems,  NKD~\cite{liu-etal-2018-knowledge} and POKS~\cite{lian2019learning}.  NKD uses a neural knowledge diffusion module  
to introduce relative entities into dialogue generation. PostKS employs both prior and posterior distributions over knowledge to select the appropriate knowledge in response generation. Then we introduce several baselines induced from our GEML framework.

	\begin{figure*}[t]
	\centering
	\includegraphics[width=0.8\linewidth]{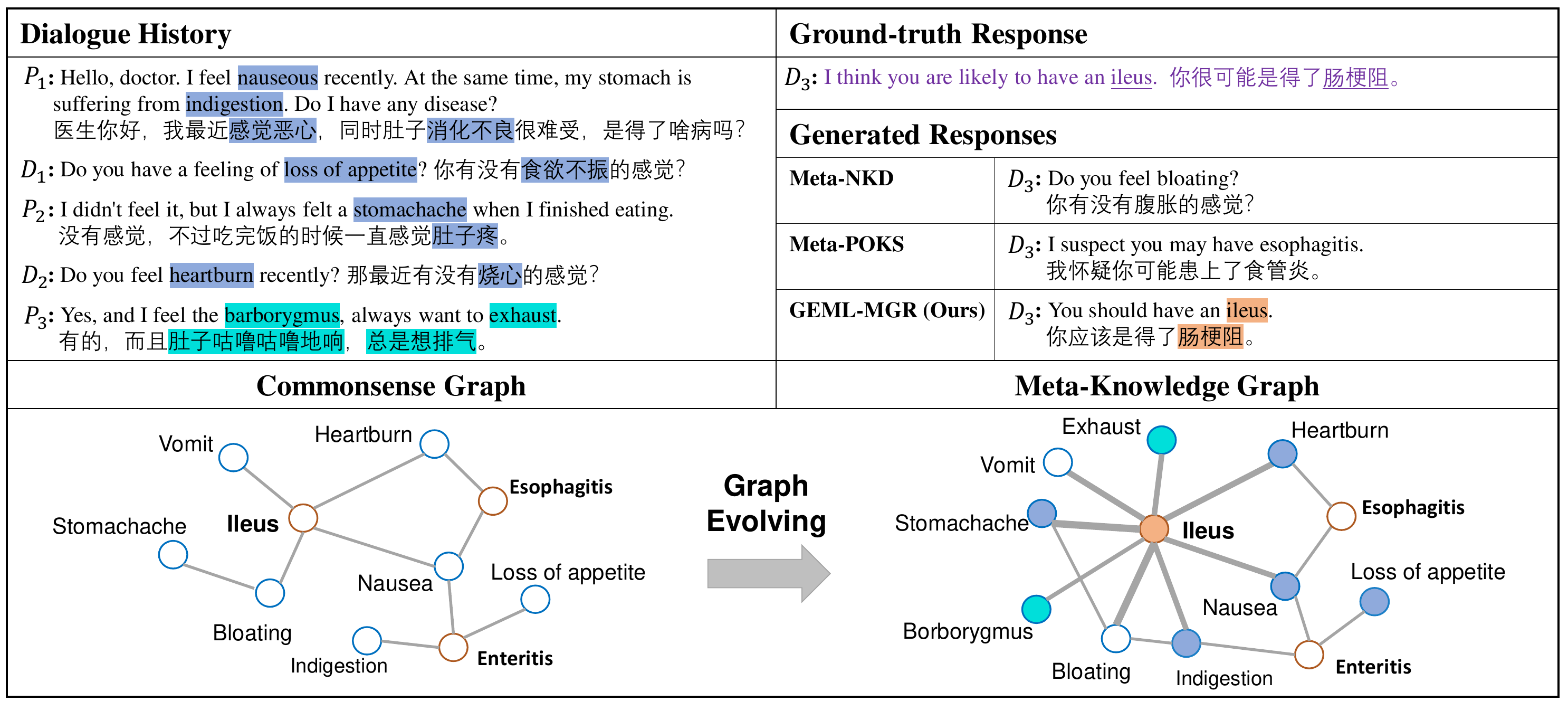}
	\caption{The visualization of the evolved meta-knowledge graph and examples of generated responses on  Chunyu. Our  graph evolving can enrich the commonsense graph and generates the response containing the correct entity (i.e., \emph{ileus}).}
	\label{fig:evolve}
	\end{figure*} 

\begin{itemize}[leftmargin=*,noitemsep]
	\item{\textit{Pre-train Only.}}  We first pre-train each dialogue base model $f_\theta$ with source diseases data in a multi-task learning paradigm, and then test it directly on target diseases. We test the above three base models and denote them as  {PT-NKD}~\cite{liu-etal-2018-knowledge},   {PT-POKS}~\cite{lian2019learning} and   {PT-MGR}  in Sec.~\ref{endtoendmodel}. This is a zero-shot learning scenario.

	\item{\textit{Fine-tuning.}} We pre-train $f_\theta$ on source diseases with the same multi-task learning paradigm and then fine-tune these pre-trained models on each target disease, which are denoted as {FT-NKD}, {FT-POKS} and {FT-MGR}.
	
	\item{\textit{Meta Learning.}}  We first meta-train three base dialogue models over source diseases with the effective meta-learning method, Reptile~\cite{nichol2018first}, and then adapt the derived meta-learners to each target disease via fine-tuning. The resulting models are denoted as  {Meta-NKD},  {Meta-POKS} and  {Meta-MGR}. 
	
	\item{\textit{GEML-MGR.}} We also employ our GEML framework on the proposed MGR model and denote it as  {GEML-MGR}. 
\end{itemize}

\begin{table}[h]
\centering
\scalebox{0.7}{
\begin{tabular}{cccc|cc|cc}
\cmidrule[\heavyrulewidth]{1-8}
\textbf{\small Graph} & \textbf{{\small Copy}} & \textbf{\small Meta-} & \textbf{\small Graph} & \multicolumn{2}{c|}{\textbf{\small CMDD}} & \multicolumn{2}{c}{\textbf{\small Chunyu}} \\
\textbf{\small Reasoning} & \textbf{{\small Mecha.}} & \textbf{\small Transfer} & \textbf{\small Evolving} & \textbf{ \small BLEU} & \textbf{{ \small Enti.-F1}} & \textbf{ \small BLEU} & \textbf{{ \small Enti.-F1}} 
 \\ \cmidrule{1-8}
\textbf{\checkmark}  & \textbf{\checkmark}   & \textbf{\checkmark} & \textbf{\checkmark} & \textbf{40.73}  &  \textbf{39.37}  & \textbf{23.07} & \textbf{38.67} \\          
                     & \textbf{\checkmark}   & \textbf{\checkmark} & \textbf{\checkmark} & 29.73  &  32.89  & 19.53 & 28.18 \\
\textbf{\checkmark}  &                      & \textbf{\checkmark}  & \textbf{\checkmark} & 32.59  &  33.78  & 21.95 &  30.81    \\                     
\textbf{\checkmark}  & \textbf{\checkmark}  &                      & \textbf{\checkmark} &  37.27 &   34.26 & 20.98 & 29.97 \\
\textbf{\checkmark}   & \textbf{\checkmark}   &   \textbf{\checkmark} &                  &  38.1  &   35.87 & 21.79 & 33.52 \\ 
\cmidrule[\heavyrulewidth]{1-8}
\end{tabular}
}
\caption{Results of ablation studies on two datasets ($\times 10^2$). }
\label{tab:ablation}
\end{table}

	\subsection{Evaluation Results}
	\textbf{Automatic Evaluation. }  
	We adopt two automatic metrics  for performance comparisons as shown in Table \ref{tab:cmdd}. 
	To evaluate the generation quality, we utilize the average of the sentence-level BLEU-1, 2, 3 and 4~\cite{chen2014systematic} and denote it as BLEU. To evaluate the success rate in the entity prediction task, we adopt Entity-F1, namely the F1 score between predicted entities in generated response and the ground-truth entities. 
	For the CMDD dataset, comparing to two other base models (NKD and POKS), our MGR always achieves the best performance in terms of both automatic and human evaluation, indicating the superiority of our end-to-end medical dialog model. The \textit{Fine-tuning} method exceeds the \textit{Pretrain-Only} in most cases and \textit{Meta-Learning} methods often outperform multi-task learning in terms of BLEU slightly yet Entity-F1 significantly. This means that the Reptile algorithm can boost the capability of knowledge reasoning and transfer over diseases. By integrating our GEML method into the MGR, we can observe significant improvement for our \textit{GEML-MGR} against all baselines, especially on Entity-F1, which demonstrates the stronger knowledge reasoning ability of our model in the medical diagnosis scenario. For the Chunyu dataset, we can observe similar results, including the superiority of the proposed GEML-MGR approach. Besides, we can see that the BLEU scores of all methods in CMDD dataset are much higher than those in the Chunyu, which demonstrates the challenge of the low-resource setting. 

	\noindent\textbf{Human Evaluation. }  	We invited five well-educated graduate students majoring in medicine to score 100 generated replies for each method. For each dataset, the evaluators are requested to grade each case in terms of  “knowledge rationality” and “generation quality” independently ranging from 1 (strongly bad) to 5 (strongly good).  The right part of Table \ref{tab:cmdd}  shows that our \textit{GEML-MGR} achieves the statistically significant higher scores than \textit{Meta-MGR} (t-test, $p$ \textless 0.01) and \textit{FT-MGR} (t-test, $p$ \textless 0.005) in terms of the two aspects.

\subsection{Discussions}

\textbf{Ablation studies. }  To verify the effects of the main components of our GEML and the base dialogue model, we conducted   ablation studies on the two datasets.   Table \ref{tab:ablation} shows that  all these factors benefit our approach. Additionally, when we drop the graph reasoning module or the graph-guided copy mechanism, there were the remarkable performance degradation on both datasets, which indicates the significance of integrating these components completely.

\noindent\textbf{Case Study of Graph Evolving. }  Fig.~\ref{fig:evolve} shows the visualization of the evolved meta-knowledge graph and cases of generated responses. One can observe an significant gap between the commonsense graph and conversations in the Chunyu dataset, as the graph cannot cover all entities in the dialogue, e.g. \emph{borborygmus} and \emph{exhaust}. Through graph evolving, the learnt meta-knowledge graph is enriched with new entities and edges that can be derived from the conversation. For instance, the meta-knowledge graph absorbed new entities \emph{borborygmus} and \emph{exhaust} and enhance the edge weights among the disease node \emph{ileus} and its neighbor nodes. Additionally, for generated responses, our GEML-MGR produces the rational and fluent diagnosis response with the least dialogue turns over other methods.

	\section{Conclusion}
	In this work, we propose an end-to-end low-resource medical dialogue generation model which meta-learns a model initialization from source diseases with the ability of fast adaptation to new diseases. Moreover, we develop a Graph-Evolving Meta-Learning (GEML) framework that learns to fast evolve a meta-knowledge graph for adapting to new diseases and reasoning the disease-symptom correlations. Accordingly, our dialogue generation model enjoys the fast learning ability and can well handle low-resource medical dialogue tasks. Experiment results testify the advantages of our approach.

\section*{Acknowledgement}
	This work was supported in part by National Natural Science Foundation of China (NSFC) under Grant No.U19A2073 and No.61976233, Guangdong Province Basic and Applied Basic Research (Regional Joint Fund-Key) Grant No.2019B1515120039, Nature Science Foundation of Shenzhen Under Grant No. 2019191361, Zhijiang Lab’s Open Fund (No. 2020AA3AB14) and CSIG Young Fellow Support Fund. 

	\bibliography{aaai21}
\end{document}